\pdfoutput=1
\documentclass[11pt]{article}

\usepackage[final]{acl}

\usepackage{times}
\usepackage{latexsym}

\usepackage[T1]{fontenc}

\usepackage[utf8]{inputenc}

\usepackage{microtype}

\usepackage{inconsolata}

\usepackage{graphicx}
\usepackage{enumitem}
\usepackage{amsmath}
\usepackage{array}
\usepackage{booktabs}
\usepackage{xcolor}
\newcolumntype{L}[1]{>{\raggedright\arraybackslash}p{#1}}
\newcounter{finding}

\makeatletter
\newcommand{\apptableofcontents}{%
  \section*{Appendix Contents}%
  \@starttoc{apx}%
}

\newcommand{\startappendixcontents}{%
  \let\origaddcontentsline\addcontentsline
  \renewcommand{\addcontentsline}[3]{%
    \def\@tempa{##1}%
    \def\@tempb{toc}%
    \ifx\@tempa\@tempb
      \origaddcontentsline{apx}{##2}{##3}%
    \else
      \origaddcontentsline{##1}{##2}{##3}%
    \fi
  }%
}
\makeatother

\title{OptiArena: Can LLMs Improve Executable Algorithms under Fixed Resource Budgets?}

\author{
 \textbf{Wenjun Peng\textsuperscript{1}},
 \textbf{Xinyu Wang\textsuperscript{1,*}}
\\
 \textsuperscript{1}Adelaide University, Australia \\
 \textsuperscript{*}Corresponding author \\
 \texttt{xinyu.wang02@adelaide.edu.au}
}

\begin{document}
\maketitle
\begin{abstract}
Static QA and code-generation benchmarks only partially capture the role that
large language models (LLMs) now play as coding agents and research tools. We
introduce \textsc{OptiArena}, a budget-controlled testbed for studying whether
LLMs can improve executable game-playing algorithms through five rounds of code
edits within a fixed minimal scaffold and under bounded evaluator feedback and
fixed resource budgets. The testbed uses two
optimization regimes, surface obfuscation controls, calibrated references,
held-out/stress splits, and diagnostics for degradation and exceptional failures,
with LLM API cost reported separately from local evaluator wall-clock. The
empirical study asks three questions: whether models can close the calibrated gap
between a designated weak starter and an editable competent baseline, whether
they can refine editable competent baselines without damaging them, and whether
gains survive surface obfuscation controls. Across twelve frontier LLMs and five
games, models improve designated weak starters more consistently than they refine
editable competent baselines, with substantial variation across games and
models.
\textsc{OptiArena} provides a practical testbed for measuring bounded-resource
algorithm optimization within the five-edit, fixed-scaffold setting studied here.
Code is available at \url{https://github.com/WJ-Peng/OptiArena}.
\end{abstract}

\section{Introduction}

Large language models (LLMs) have changed what it means to evaluate progress in NLP. The field has long relied on benchmarks that ask a model for an answer to a static input, such as a multiple-choice answer, a short response, a translation, or a program from a specification~\citep{hendrycks2021measuring,srivastava2023beyond,chen2021evaluating}. These benchmarks remain useful but measure only a narrow slice of the role that LLMs now play. Recent LLMs increasingly serve not merely as answer generators but as coding agents and research assistants that read feedback, revise executable artifacts, and participate in iterative workflows. SWE-bench~\citep{jimenez2024swe} and ProgramBench~\citep{yang2026programbench} have pushed code evaluation toward realistic software engineering, from repository repair under hidden tests to reconstructing programs from executables and documentation. Together, these developments raise a sharper question: can an LLM improve an existing algorithm through bounded feedback under fixed edit and execution budgets?

This question is also relevant to the broader agenda of AI-assisted discovery, where systems such as FunSearch~\citep{romera2024mathematical}, Eureka~\citep{ma2024eureka}, The AI Scientist~\citep{lu2024ai}, AlphaEvolve~\citep{novikov2025alphaevolve}, and the AI co-mathematician~\citep{zheng2026aicomath} place LLMs in evaluator-guided search loops but are difficult to compare, since results often depend on substantial, uncontrolled search and engineering budgets. Our study instead targets a narrower, budget-explicit diagnostic setting rather than the complexity of repository-scale engineering or open-ended scientific discovery.

We introduce \textsc{OptiArena}, a budget-controlled testbed for studying LLMs as code-level algorithm optimizers. The testbed uses games as controlled optimization substrates because their rules are precise, execution is replayable, and performance can be measured through repeated interaction with hidden references. In \textsc{OptiArena}, the object being optimized is executable strategy code. An LLM receives a runnable policy through a fixed minimal scaffold, edits it over five rounds, and observes structured feedback from a sandboxed evaluator with fixed resource limits. This setup separates two capabilities that are often conflated in code evaluation. One is recovering useful behavior from designated weak starter policies, and the other is making safe local improvements to editable competent baselines. We instantiate these as weak-start optimization and strong-baseline refinement, and analyze both under held-out and stress evaluation. These shared anchors are registered design choices rather than claims of global worst-case or optimal performance, so the comparison is conditional on their construction and remaining headroom.

A final challenge is task familiarity: benchmark contamination remains a threat to LLM evaluation~\citep{deng2024investigating}. Because many classic game algorithms and heuristics are widely documented, success on familiar task surfaces may reflect recognition of standard templates rather than robust algorithmic improvement. \textsc{OptiArena} therefore makes task-surface familiarity part of the evaluation design, asking whether gains persist when recognizable cues are removed. Where an independent capability signal is available, we also use it to identify evaluator overfitting rather than treating every gain against a bounded feedback evaluator as general improvement. In summary, the contribution of this paper is threefold:

\begin{itemize}[leftmargin=*, itemsep=2pt, topsep=2pt]
    \item We propose \textsc{OptiArena}, a budget-controlled testbed that studies LLMs as bounded-resource optimizers of executable strategy code across controlled game environments.
    \item We introduce a two-regime evaluation protocol that separates improvement from designated weak starters and refinement of editable competent baselines, providing a controlled way to study code-level optimization under bounded feedback.
    \item We study 12 frontier LLMs across five game families, measuring not only final performance but also robustness to held-out evaluation and surface obfuscation, while reporting exceptional failures and separating LLM API cost from local evaluator wall-clock.
\end{itemize}

\section{Related Work}

\subsection{Code and Agent Evaluation}
Many LLM benchmarks evaluate static prediction, scoring a model against a fixed reference answer or test suite: broad language and reasoning suites~\citep{hendrycks2021measuring,srivastava2023beyond}, code-generation benchmarks such as HumanEval~\citep{chen2021evaluating}, MBPP~\citep{austin2021program}, and APPS~\citep{hendrycks2021apps}, and more recent benchmarks that add execution-oriented subtasks, repository edits, or full-program reconstruction from external artifacts~\citep{jain2025livecodebench,jimenez2024swe,yang2026programbench}. Agent benchmarks and coding-agent interfaces further move evaluation toward multi-turn interaction, tool use, and repository navigation~\citep{liu2024agentbench,yang2024sweagent}.

Executable feedback is also central to program repair and code optimization. Search-based repair systems such as GenProg~\citep{legoues2012genprog} use automated tests to modify programs, and recent LLM work studies edits that improve the runtime of functionally equivalent code~\citep{shypula2024learning}. Among these directions, game-based code evaluation is closest to our setting. ProxyWar~\citep{peng2026proxywar} embeds LLM-generated agents in competitive arenas and evaluates them through tests, repair loops, and tournament outcomes. \textsc{OptiArena} also uses games for executable evaluation, but the model edits existing strategy code over bounded feedback rounds. The main outcome is therefore algorithmic improvement and baseline degradation, rather than general code-generation quality, pass/fail correctness, or raw runtime speed.

\subsection{LLM-Guided Optimization and Discovery}
Recent systems increasingly combine LLMs with automated evaluators, archives, or evolutionary search. OPRO~\citep{yang2024large} uses LLMs as black-box optimizers over natural language descriptions; Eureka~\citep{ma2024eureka} evolves reward code for reinforcement learning; FunSearch~\citep{romera2024mathematical} and Evolution of Heuristics~\citep{liu2024evolution} search over executable programs and heuristics; The AI Scientist~\citep{lu2024ai} and AlphaEvolve~\citep{novikov2025alphaevolve} broaden the loop to scientific artifacts and algorithmic discovery. Parallel work in non-LLM algorithm discovery, such as AlphaTensor~\citep{fawzi2022discovering} and AlphaDev~\citep{mankowitz2023faster}, also formulates algorithm search as a game with programmatic feedback. Closer to the edit-feedback loop, recent heuristic learning~\citep{weng2026learning_beyond_gradients} iteratively revises executable RL policies from logs, tests, replays, and environment feedback, and the Automated LLM Speedrunning Benchmark~\citep{zhao2025speedrunning} tests whether agents can reproduce NanoGPT improvements under resource constraints and strong scaffolds. Rather than proposing another discovery engine or scaffold, \textsc{OptiArena} fixes a minimal scaffold, the feedback channel, and resource limits, and withholds the scoring reference, so that differences reflect a model's ability to revise executable algorithms under bounded feedback.

\subsection{Game-Based Evaluation}
Games have long served as controlled environments for measuring general decision-making systems, from Atari to OpenSpiel's collection of single-agent and multi-agent games~\citep{bellemare2013arcade,lanctot2019openspiel}. Their methodological appeal is that rules are precise, episodes are replayable, and performance can be measured over many trials with calibrated opponents or oracles. \textsc{OptiArena} uses these properties for LLM evaluation rather than for training a game-playing agent. The model edits source code for a policy, and the evaluator runs that code in a sandbox against hidden references, held-out openings, and stress splits.

Recent game-centered work studies interactive reasoning~\citep{shi2025korgym}, probabilistic decision making~\citep{peng2026poker}, explanations in imperfect-information settings~\citep{wang-etal-2026-triex}, and vision-language game-play evaluation~\citep{wang2025large}. \textsc{OptiArena} instead evaluates code-level revision of a persistent policy across sealed games under a fixed edit budget.

Contamination-aware benchmarks often rely on temporal freshness, updated
question pools, or newly collected programming tasks
~\citep{jain2025livecodebench,white2025livebench}. This strategy is less
straightforward for algorithmic games: the rules are intentionally stable, and
many strong ideas, from minimax and alpha-beta pruning to game-specific
heuristics, have circulated for decades. \textsc{OptiArena} therefore
complements freshness-based evaluation with surface-based interventions. It
keeps final evaluation hidden, enforces the same per-move resource limits
across submissions, and compares familiar tasks with obfuscated task surfaces.
This design tests whether performance gains come from executable strategy
improvements rather than from recognizing a known game or spending more search
time.

\section{OptiArena}
\label{sec:method}
\label{sec:optiarena-overview}

\textsc{OptiArena} casts algorithm improvement as a controlled code-editing problem. An evaluation instance binds a model to a predeclared task condition and runs a single bounded loop of editing, evaluation, and candidate selection. The model receives five opportunities to revise a runnable policy from sandboxed development feedback, and after the five-edit budget one candidate is selected for sealed held-out and stress evaluation.

Two properties keep the setting controlled. First, it is \emph{sealed}: the model sees development feedback but never the hidden reference that anchors the score, nor the final evaluation material. We deliberately separate two objects that are easy to conflate. The \emph{editable competent baseline} \(\pi_{\mathrm{comp}}^g\) is source code the model can read and modify. The \emph{hidden reference} is a fixed-budget opponent, engine, or reference policy that scores candidates and is never exposed. The editable baseline supplies a calibration anchor for WSO where GCR is used, as well as the starting point for refinement, while the hidden reference supplies the score. Second, every candidate runs under the \emph{same} registered resource policy as its starter, so a candidate that exceeds its registered compute budget is counted as a failure rather than as progress.

The benchmark separates three capabilities that static code benchmarks tend to conflate, which gives three research questions.

\noindent\textbf{RQ1: Weak-start optimization.} Under a fixed resource budget and evaluator-provided development feedback, can LLMs improve designated weak starters, and how much registered capability do they recover?

\noindent\textbf{RQ2: Strong-baseline refinement.} When given an editable competent baseline, do models preserve, improve, or damage the algorithm under the same resource policy?

\noindent\textbf{RQ3: Contamination sensitivity.} Do weak-start capability gains depend on recognizable task-surface cues, or do they survive semantics-preserving obfuscation of the task surface?

\subsection{Task Suite and Policy Interface}
\label{sec:task-suite}

The task suite contains five compact game families (Table~\ref{tab:benchmark-tasks}). Four are two-player zero-sum games, while 2048 is a stochastic single-player planning task. Their value comes from controllability rather than human novelty, since each environment supports repeated execution against calibrated references under a fixed budget. The suite is intentionally small enough for repeated held-out evaluation while still spanning distinct algorithmic demands.

\begin{table}[!ht]
\centering
\footnotesize
\setlength{\tabcolsep}{3pt}
\renewcommand{\arraystretch}{1.08}
\begin{tabular}{@{}L{0.30\columnwidth}L{0.22\columnwidth}L{0.38\columnwidth}@{}}
\toprule
Family & Board & Signal \\
\midrule
Breakthrough & 8$\times$8 & WR; MA \\
Othello & 8$\times$8 & MA; WR \\
Hex & 11$\times$11 & MA; WR \\
Connect 4 & 7$\times$6 & OA; regret \\
2048 & 4$\times$4 & Score; tile rate \\
\bottomrule
\end{tabular}
\caption{\textbf{Benchmark task suite.} Signals are listed primary first (e.g.\ ``MA; WR'' means move agreement is primary, win rate secondary); WR denotes win rate, MA move agreement, and OA oracle agreement.}
\label{tab:benchmark-tasks}
\end{table}

Across tasks, the model edits a reusable policy module rather than issuing individual moves during evaluation. Canonical rules are backed by OpenSpiel where available.
The harness handles game execution and sandbox bookkeeping, so the prompt-exposed strategy code is the only object the model can change. This keeps the unit of optimization clear while preventing privileged access to the evaluator or external tools. The exact interface is specified in Appendix~\ref{app:policy-contract}.

\subsection{Optimization Regimes and Feedback Loop}
\label{sec:optimization-protocol}

Each evaluation instance runs the same loop of editing, evaluation, and candidate selection. Iteration 0 evaluates the unmodified starter and initializes the archive. For each of five subsequent edit calls, the harness selects an archived parent by \(\epsilon\)-greedy selection with \(\epsilon=0.2\) and prompts the LLM with the parent source, task rules, state encoding, and compact development feedback; in the primary \texttt{trace\_aware} condition the model also reads short execution traces for representative games. The feedback summarizes validity (syntax, interface, illegal moves, crashes, timeouts), performance (win rate or robust score with confidence intervals), and efficiency (game length and move time), but it never exposes the hidden-reference source, held-out openings, or solver values. The model returns a complete Python source file, which the harness turns into an auditable evaluation record; invalid edit attempts are retained in the optimization archive rather than silently repaired.

The two regimes differ only in the editable source exposed to the optimizer. In \textit{Weak-Start Optimization} (WSO), the designated weak starter \(\pi_0^g\) is deliberately weak so that there is calibrated headroom for improvement. The target is not to beat the strongest available engine, but to recover capability from the designated weak starter under the registered per-game estimator. In \textit{Strong-Baseline Refinement} (SBR), the starter is an editable competent policy \(\pi_{\mathrm{comp}}^g\). Here, ``strong'' is relative to the designated weak starter and does not imply global optimality. This regime asks whether the model can make local improvements while preserving the algorithmic structure and resource discipline that make the baseline competent under the registered budget.

After the five edit calls are exhausted, a fixed development rule selects one candidate for final development, held-out, and stress evaluation. The rule is fixed before the run and uses only development feedback. The archive retains the complete candidate lineage and evaluation metadata for audit. Appendix~\ref{app:optimization-loop} gives the full loop and feedback schema.

\subsection{Sealed Splits and Surface Controls}
\label{sec:evaluation-controls}

\paragraph{Resource policy.} Candidates execute in an isolated Docker sandbox with one CPU, 2\,GB memory, a 64-process limit, disabled network access, and hard per-move timeouts. The default budget is 1.0\,s per move for the two-player games and 0.1\,s per move for 2048, with per-game wall-clock caps that prevent hanging evaluations. These budgets are archived as evaluation metadata.

\paragraph{Splits.} The benchmark uses development, held-out, and stress splits. Development feedback guides optimization; held-out and stress rows are produced only after candidate selection. Held-out uses disjoint openings or seed pools, while stress keeps the candidate resource policy fixed and shifts to a harder registered distribution. The hidden reference that scores these splits is a fixed-budget opponent or engine for the two-player games and a stronger reference policy for 2048.

\paragraph{Surface control.} For WSO, the surface control compares a \textit{vanilla} surface against an \textit{obfuscated} one. The vanilla condition uses standard game terminology and source identifiers; the obfuscated condition rewrites the prompt-visible surface while preserving game dynamics exactly. This isolates sensitivity to familiar cues without changing the underlying task.

\subsection{Metrics}
\label{sec:metrics}

\textsc{OptiArena} assigns one primary quantity to each research question: WSO uses weak-start capability, SBR uses improvement over the editable competent baseline, and the contamination analysis uses WSO capability sensitivity to surface familiarity. Validity, degradation, and cost are diagnostic context rather than separate questions.

Let \(P_g(\pi)\) denote the registered primary performance of policy \(\pi\) on game \(g\), computed only from calibrated or final-evaluation rows. For two-player games \(P_g\) is typically win rate or point score; for 2048 it is a robust score over a fixed seed pool. The three primary metrics are all defined on top of \(P_g\).

\paragraph{WSO capability.} Because several games are scored against very strong references, raw win rate can saturate at zero for both the designated weak starter and the editable competent baseline, which makes a win-rate ratio uninformative. We therefore report a single \emph{capability estimator} (Cap) per game, registered in advance according to the strongest reference signal each game exposes: a move-optimal rate against a solved-game oracle (Connect~4), a move-agreement rate against Edax for Othello and MoHex for Hex, a naive-to-reference normalized score for the stochastic single-player task (2048), and, when the gap is well conditioned, the gap-closing ratio (Breakthrough)
\begin{equation}
\mathrm{GCR}(g,m)=
\frac{P_g(\pi_T^{g,m}) - P_g(\pi_0^g)}
     {P_g(\pi_{\mathrm{comp}}^g) - P_g(\pi_0^g)},
\label{eq:gcr}
\end{equation}
whose denominator comes from calibration runs of the designated weak starter and editable competent baseline against the same hidden reference. GCR is reported unclipped: values below \(0\) indicate degradation relative to the designated weak starter, and values above \(1\) indicate surpassing the calibrated competent baseline. Appendix~\ref{app:saturated-games} lists the per-game estimator and its fallbacks.

\paragraph{SBR improvement.} For SBR we report the change from the editable competent baseline under matched evaluation material and the same resource policy,
\begin{equation}
\Delta_{\mathrm{SBR}}(g,m)=P_g(\pi_T^{g,m})-P_g(\pi_{\mathrm{comp}}^g),
\end{equation}
so positive values mean the candidate improves on the editable competent baseline; for 2048, \(P_g\) is the registered robust score and the same definition applies.

\paragraph{Contamination sensitivity.} For the WSO surface control, we compare the vanilla and obfuscated conditions using the same registered per-game Cap estimator; in the definition below, \(M=\mathrm{Cap}\) and \(c=\mathrm{obfuscated}\):
\begin{equation}
\mathrm{CSI}_{c}(g,m)=M_{\mathrm{vanilla}}(g,m)-M_{c}(g,m).
\end{equation}
Positive CSI indicates an advantage for the familiar vanilla surface; a small CSI together with positive absolute performance is stronger evidence that the model optimizes from feedback rather than recalling a known algorithm. Both conditions must resolve to the same Cap estimator; otherwise the comparison is reported as unavailable rather than subtracting incompatible quantities. Cross-game aggregates use normalized capability metrics rather than raw game scores (Appendix~\ref{app:metric-diagnostics}).

\paragraph{Diagnostics.} Validity is the fraction of attempted final-evaluation trials that complete without invalid behavior or candidate timeout; degradation is the fraction of cells for which a model's selected candidate scores below its starter on a game; cost is the optimization-loop LLM cost from provider usage metadata, kept separate from local evaluation time. Candidate-caused trial failures during final evaluation remain benchmark outcomes and lower validity; only infrastructure-class failures permit replacement, with the original attempt retained. Final comparisons use the same selected candidate on paired registered opening or seed material with reported sample sizes, and aggregate comparisons resample over matched model-game cells (bootstrap) rather than treating missing rows as neutral outcomes.

\section{Experiments}
\label{sec:experiments}

The experiments instantiate the benchmark around the three questions introduced
in Section~\ref{sec:optiarena-overview}. Rather than treating the benchmark as
a collection of engineering checks, we read the results as evidence about three
forms of optimization behavior: whether feedback helps models repair weak code,
whether models can improve competent code without damaging it, and whether
observed gains survive when familiar game surfaces are weakened.

\subsection{Experimental Setup}
\label{sec:exp-protocol}

We evaluate twelve LLMs through the same code-editing interface, with no tools,
no web access, and no access to evaluator internals. The proprietary/API roster
contains Claude Sonnet 4.6~\citep{anthropic2026systemcards}, GPT-5.4 and
GPT-5.1-Codex~\citep{openai2026gpt54,openai2026gpt51codex}, Gemini 3.5
Flash~\citep{google2026gemini35flash}, MiniMax-M2.7~\citep{minimax2026m27},
and Kimi-K2.6~\citep{moonshot2026kimi}. The open-source/open-weight roster
contains DeepSeek-V4-Pro~\citep{deepseek2026v4}, Qwen3-Coder and
Qwen3-14B~\citep{qwen2025coder,qwen2025qwen3},
Gemma-4-26B-A4B-IT (Gemma-4-26B)~\citep{google2026gemma4},
Mistral-Small-2603 (Mistral-Small)~\citep{mistral2026small}, and
GLM-4.7~\citep{zai2025glm47}. Exact provider ids, snapshots, decoding
settings, and API dates are stored in run metadata.
Each run gives the model five edit calls after iteration 0, which evaluates the unmodified starter. Intermediate
candidates receive three-game quick evaluations, and the selected candidate is then
evaluated on development, held-out, and stress splits with \(n=50\) games or
trials per split. Candidate policies receive 1.0 second per move in two-player
games and 0.1 seconds per move in 2048. The main matrices use seed 0 and registered, disjoint opening or seed pools; two-player aggregate counts use Wilson intervals where trial-level pairing is unavailable. Thus each cell for a model, game, and regime
uses a bounded number of LLM edit calls and a fixed evaluator budget; token
usage and provider-reported cost are archived separately from local
game-evaluation time.

The reporting policy fixes coverage before results are analyzed. The first
matrix reports WSO on the vanilla condition across the full model-game roster;
the second reports the matched SBR matrix; and the third reports the
WSO obfuscated condition for contamination sensitivity. Each main matrix contains 60 model--game cells and 180 final split records. Aggregate claims are
made only for matrices that satisfy the declared coverage and metadata checks;
partial runs are reported separately. Appendix~\ref{app:run-config} and
Appendix~\ref{app:experimental-protocol} give the optimization budget, model
roster, inclusion rules, and artifact metadata policy.

\subsection{Optimization Results: Weak Starts and Competent Baselines}
\label{sec:quant-results}

Table~\ref{tab:rq-optimization} summarizes both regimes at the model level; the
per-game tables (Tables~\ref{tab:wso-full} and~\ref{tab:sbr-full}) are the audit
surface for heterogeneity, saturation, and game-specific failures. All entries
come from archived calibration and final-evaluation rows; invalid programs,
illegal moves, and timeouts remain benchmark outcomes rather than missing data.

\begin{table}[t]
\centering
\small
\setlength{\tabcolsep}{4pt}
\resizebox{\columnwidth}{!}{%
\begin{tabular}{@{}lcccc@{}}
\toprule
Model & WSO & SBR & Deg. & Cost \\
\midrule
\multicolumn{5}{@{}l}{\textit{Proprietary}} \\
Claude Sonnet 4.6 & 0.37 & $-$0.04 & 0.40 & \$0.072 \\
GPT-5.4 & 0.50 & $-$0.04 & 0.60 & \$0.074 \\
GPT-5.1-Codex & 0.56 & $-$0.02 & 0.40 & \$0.180 \\
Gemini 3.5 Flash & \textbf{0.67} & \textbf{$+$0.12} & \textbf{0.00} & \$0.258 \\
\midrule
\multicolumn{5}{@{}l}{\textit{Open-source / open-weight}} \\
DeepSeek-V4-Pro & 0.50 & $+$0.06 & 0.20 & \$0.016 \\
Qwen3-Coder & 0.33 & $-$0.01 & 0.40 & \$0.006 \\
Qwen3-14B & 0.35 & $-$0.03 & 0.20 & \$0.001 \\
Gemma-4-26B & 0.61 & $-$0.02 & 0.40 & \$0.001 \\
MiniMax-M2.7 & 0.52 & $+$0.03 & 0.40 & \$0.012 \\
Kimi-K2.6 & 0.41 & $+$0.05 & \textbf{0.00} & \$0.027 \\
Mistral-Small & 0.31 & $-$0.01 & 0.40 & \$0.002 \\
GLM-4.7 & 0.52 & $-$0.05 & 0.40 & \$0.018 \\
\bottomrule
\end{tabular}}
\caption{\textbf{Per-model optimization summary.} WSO is the mean registered weak-start
capability estimator (Cap) across games; SBR is the mean
\(\Delta_{\mathrm{SBR}}\) relative to the editable competent baseline. Deg.\
(degradation rate) and Cost (mean optimization-loop LLM cost per edit round,
USD) are reported for the SBR regime; Per-game matrices appear in
Tables~\ref{tab:wso-full} and~\ref{tab:sbr-full}.}
\label{tab:rq-optimization}
\end{table}

\noindent\textbf{Weak-start optimization.}
Table~\ref{tab:wso-full} keeps per-game capability and the development-to-held-out
and held-out-to-stress gaps visible, so a development-only gain that vanishes under
held-out evaluation is not credited.

Weak-start capability remains strongly model- and game-dependent: mean
Cap ranges from 0.31 to 0.67, led by Gemini 3.5 Flash (0.67), Gemma-4-26B
(0.61), and GPT-5.1-Codex (0.56). Across models, mean Cap is highest on
Connect~4 (0.82) and 2048 (0.80), while agreement with MoHex on Hex remains
low (mean 0.04). Five models exceed the 2048 reference and two exceed the
competent-baseline anchor on Breakthrough. These gains do not imply uniform
improvement: only 57 of 60 selected WSO candidates complete all 50 held-out
trials, and Connect~4 win-rate changes sometimes diverge from solved-oracle
optimality (Section~\ref{sec:diagnostic-analysis}).

\begin{table*}[t]
\centering
\footnotesize
\setlength{\tabcolsep}{2.5pt}
\begin{tabular}{@{}L{0.18\textwidth}cccccccc@{}}
\toprule
Model & BT & C4 & Oth & Hex & 2048 & Mean Cap & Dev/HO & HO/Stress \\
\midrule
\multicolumn{9}{@{}l}{\textit{Proprietary models}} \\
Claude Sonnet 4.6 & 0.00/0 & 0.81/$-$ & \textbf{0.44}/$+$ & 0.07/$+$ & 0.52/$+$ & 0.37 & $+$0.10 & $-$0.10 \\
GPT-5.4 & 0.00/0 & \textbf{0.92}/$+$ & 0.40/0 & 0.03/0 & 1.14/$+$ & 0.50 & $-$0.14 & $+$0.18 \\
GPT-5.1-Codex & 0.33/$+$ & 0.81/0 & 0.40/0 & 0.03/0 & 1.22/$+$ & 0.56 & $+$0.04 & $-$0.08 \\
Gemini 3.5 Flash & \textbf{1.33}/$+$ & 0.91/$+$ & 0.40/0 & 0.03/0 & 0.68/$+$ & \textbf{0.67} & $-$0.10 & $+$0.32 \\
\midrule
\multicolumn{9}{@{}l}{\textit{Open-source / open-weight models}} \\
DeepSeek-V4-Pro & 0.00/0 & 0.81/0 & 0.40/0 & \textbf{0.09}/0 & 1.21/$+$ & 0.50 & $-$0.10 & $+$0.06 \\
Qwen3-Coder & 0.00/0 & 0.81/0 & 0.33/$-$ & 0.00/0 & 0.51/$+$ & 0.33 & $+$0.02 & $-$0.10 \\
Qwen3-14B & 0.67/$+$ & 0.81/0 & 0.40/0 & 0.03/0 & $-$0.15/$-$ & 0.35 & $+$0.10 & $-$0.08 \\
Gemma-4-26B & \textbf{1.33}/$+$ & 0.81/0 & 0.35/0 & 0.03/0 & 0.52/$+$ & 0.61 & $+$0.04 & $-$0.08 \\
MiniMax-M2.7 & 0.00/0 & 0.81/0 & 0.41/0 & 0.02/$+$ & 1.35/$+$ & 0.52 & $+$0.04 & $-$0.08 \\
Kimi-K2.6 & 0.00/0 & 0.81/0 & 0.40/$+$ & 0.03/0 & 0.82/$+$ & 0.41 & $+$0.04 & $-$0.08 \\
Mistral-Small & 0.00/0 & 0.81/0 & 0.35/0 & 0.03/0 & 0.36/$+$ & 0.31 & $+$0.04 & $-$0.08 \\
GLM-4.7 & 0.00/0 & 0.81/0 & 0.40/0 & 0.03/0 & \textbf{1.39}/$+$ & 0.52 & $-$0.08 & $+$0.02 \\
\bottomrule
\end{tabular}
\caption{\textbf{Full WSO held-out table.} Each cell reports Cap/$\Delta$, where $\Delta\in\{+,-,0\}$ is the sign of the change from the starter on the capability scale; per-game capability estimators are C4 = move-optimal rate vs solved-game BitBully, BT = GCR, Othello/Hex = move-agreement rate vs Edax/MoHex, 2048 = reference-$\Delta$ on robust score; Mean Cap averages the available per-game capabilities. Dev/HO and HO/Stress are the development-to-held-out and held-out-to-stress win-rate gaps on Connect4, shown as a representative robustness diagnostic (positive Dev/HO indicates held-out drop; per-game splits are archived). Validity (the completed-trial fraction free of invalid behavior, illegal moves, or timeout) is 1.00 in every cell except Claude Sonnet 4.6 (2048, 0.88), DeepSeek-V4-Pro (Hex, 0.62; 2048, 0.96).}
\label{tab:wso-full}
\end{table*}

\noindent\textbf{Competent-baseline refinement.}
SBR is harder: a useful model must improve a competent baseline while preserving
the structure that makes it competent under the registered budget. Negative values and high degradation are
therefore substantive outcomes because they reveal destructive rewrites, broken
invariants, or search that fails under the time budget. Table~\ref{tab:sbr-full}
reports the per-game surface behind the aggregate column.

Refinement is markedly harder and more game-dependent than weak-start
repair. On the four two-player games, Gemini 3.5 Flash has the largest mean
improvement ($+0.12$), followed by DeepSeek-V4-Pro ($+0.06$); seven models have
non-positive mean change. Breakthrough supplies most positive refinements,
whereas ten of twelve models degrade on Connect~4 and all models are neutral on
the saturated Othello and Hex win-rate surfaces. On 2048, seven of twelve models improve over the competent baseline on the
registered normalized SBR score, four degrade it, and one is unchanged, a more
mixed picture than the two-player surfaces. Fifty-eight of
60 selected SBR candidates complete all held-out trials, so these negative and
neutral outcomes cannot be explained by treating candidate failures as missing
data.

\begin{table*}[t]
\centering
\footnotesize
\setlength{\tabcolsep}{3pt}
\begin{tabular}{@{}L{0.18\textwidth}cccccccc@{}}
\toprule
Model & BT & C4 & Oth & Hex & 2048 & Mean \(\Delta\) & Deg. & Cost \\
\midrule
\multicolumn{9}{@{}l}{\textit{Proprietary models}} \\
Claude Sonnet 4.6 & $-$0.06 & $-$0.08 & 0.00 & 0.00 & $+$0.05 & $-$0.04 & 0.40 & \$0.072 \\
GPT-5.4 & $-$0.08 & $-$0.08 & 0.00 & 0.00 & $-$0.04 & $-$0.04 & 0.60 & \$0.074 \\
GPT-5.1-Codex & 0.00 & $-$0.08 & 0.00 & 0.00 & $+$0.03 & $-$0.02 & 0.40 & \$0.180 \\
Gemini 3.5 Flash & $+$0.12 & \textbf{$+$0.36} & 0.00 & 0.00 & $+$0.23 & \textbf{$+$0.12} & \textbf{0.00} & \$0.258 \\
\midrule
\multicolumn{9}{@{}l}{\textit{Open-source / open-weight models}} \\
DeepSeek-V4-Pro & \textbf{$+$0.34} & $-$0.12 & 0.00 & 0.00 & $+$0.37 & $+$0.06 & 0.20 & \$0.016 \\
Qwen3-Coder & $+$0.04 & $-$0.06 & 0.00 & 0.00 & $-$0.44 & $-$0.01 & 0.40 & \$0.006 \\
Qwen3-14B & $+$0.04 & $-$0.16 & 0.00 & 0.00 & \textbf{$+$0.50} & $-$0.03 & 0.20 & \$0.001 \\
Gemma-4-26B & $+$0.04 & $-$0.12 & 0.00 & 0.00 & $-$0.13 & $-$0.02 & 0.40 & \$0.001 \\
MiniMax-M2.7 & $+$0.18 & $-$0.08 & 0.00 & 0.00 & $-$0.02 & $+$0.03 & 0.40 & \$0.012 \\
Kimi-K2.6 & $-$0.02 & $+$0.20 & 0.00 & 0.00 & $+$0.20 & $+$0.05 & \textbf{0.00} & \$0.027 \\
Mistral-Small & 0.00 & $-$0.04 & 0.00 & 0.00 & $+$0.12 & $-$0.01 & 0.40 & \$0.002 \\
GLM-4.7 & $-$0.02 & $-$0.16 & 0.00 & 0.00 & $+$0.00 & $-$0.05 & 0.40 & \$0.018 \\
\bottomrule
\end{tabular}
\caption{\textbf{Full SBR held-out table.} Task entries report \(\Delta_{\mathrm{SBR}}\)
relative to the empirical bounded-resource competent baseline; Mean \(\Delta\)
averages the two-player games (same win-rate scale), with 2048 reported
separately as a robust-score delta. Deg. is the degradation rate; Cost is the
mean optimization-loop LLM cost per edit round (USD).}
\label{tab:sbr-full}
\end{table*}

\subsection{Contamination Sensitivity}
\label{sec:contamination-results}

The third question asks whether optimization survives when recognizable surface
cues are weakened. The matched WSO control compares vanilla performance against
the obfuscated condition. A large positive CSI indicates that a
model performs better on the familiar surface than on the comparison condition;
a small CSI with positive absolute performance provides stronger evidence that
the model is using feedback to optimize rather than simply recalling a known
game algorithm.

\begin{table}[t]
\centering
\small
\setlength{\tabcolsep}{5pt}
\begin{tabular}{@{}lrr@{}}
\toprule
Scope & Mean CSI-Cap & 95\% CI \\
\midrule
Overall & $-0.043$ & $[-0.124,\ 0.043]$ \\
Breakthrough & $+0.139$ & $[-0.083,\ 0.417]$ \\
Connect~4 & $+0.003$ & $[-0.020,\ 0.028]$ \\
Hex & $-0.001$ & $[-0.037,\ 0.033]$ \\
Othello & $-0.059$ & $[-0.102,\ -0.020]$ \\
2048 & $-0.298$ & $[-0.560,\ -0.011]$ \\
\bottomrule
\end{tabular}
\caption{\textbf{WSO contamination sensitivity.} CSI-Cap is vanilla Cap minus
obfuscated Cap. Each game contains 12 matched model pairs; the overall row
contains all 60 model--game pairs. Intervals use 10,000 paired bootstrap
resamples over the matched cells.}
\label{tab:rq3-csi}
\end{table}

The cross-game mean CSI-Cap is $-0.043$ (95\% CI
$[-0.124, 0.043]$). The scale-normalized aggregate is likewise close to zero
at $0.016$ (95\% CI $[-0.105, 0.144]$). Thus the control provides no evidence
of an aggregate advantage for the familiar surface, while the per-game signs
are heterogeneous: Connect~4 and Hex are near zero, Othello and 2048 favor the
obfuscated surface, and Breakthrough has a positive but imprecise shift.

\subsection{Diagnostic and Qualitative Analysis}
\label{sec:diagnostic-analysis}

The aggregate numbers are interpreted together with diagnostics rather than in
isolation. Validity identifies whether a model can keep generated code
executable. Development-to-held-out and held-out-to-stress gaps measure whether
optimization transfers beyond the feedback split. Degradation rate separates
small positive edits from brittle rewrites, and cost distinguishes efficient
optimization from gains purchased by many expensive attempts.

The archive also supports qualitative auditing: cases such as a successful
weak-to-strong repair, a destructive SBR edit, an invalid program, or a timeout
from excessive search are selected by explicit rules over archived runs and
released with the artifact (Appendix~\ref{app:qualitative}), so the mechanisms
behind the aggregate numbers can be inspected rather than asserted.

Across the 60 WSO cells, the selected candidate keeps the starter
in 6 cells, inserts a known algorithm in 24, and expands heuristics without a
known-algorithm insertion in 30. Across the 60 SBR cells, the selected candidate keeps the
starter in 47 cells, applies a destructive edit in 5 (four of which still
insert a known algorithm), and applies a positive edit in 8 (two via algorithm
insertion, six via other optimization); no cell is neutral. As with the WSO
census, these archive-derived categories describe the selected edit's
direction and mechanism rather than its quality. Table~\ref{tab:qualitative-cases}
(Appendix~\ref{app:qualitative}) lists four rule-selected diffs illustrating
these counts: the two largest algorithm-inserted WSO Cap gains (a single-player
and a two-player game), the most negative $\Delta_{\mathrm{SBR}}$ (a
from-scratch rewrite of an already-working search that regresses), and the
most positive $\Delta_{\mathrm{SBR}}$ (an in-place fix to the existing
evaluation function).

\noindent\textbf{Win rate and move optimality can diverge.}
On Connect~4 the two natural weak-start signals do not move together uniformly. Five models raise their win rate against the depth-bounded
BitBully reference; among them, one lowers solved-oracle move optimality and two
leave it unchanged. We
read this as a behavioral observation rather than a metric artifact. The edited
policies win more games without playing more game-theoretically optimal moves,
for instance by exploiting the bounded-depth reference rather than approximating
perfect play. The main WSO table therefore reports the solved-game optimality
estimator, which is anchored to ground truth and does not credit such
exploitation; we surface the divergence here so that a higher win rate is not
read as uniformly stronger play. We make no normative claim about which behavior
is preferable, only that the two signals measure different things.
As a supplementary check on the primary five-edit budget, we run
a 10- and 15-edit extension on all four games studied here (vanilla
condition only). On the three games with a stable
capability estimator, median WSO Cap shows no consistent trend across the
5/10/15-edit budgets (0.74/0.72/0.72). Median SBR improvement over the
competent baseline likewise stays near zero at every budget
(0.00/0.00/$-$0.02), with one severe destructive edit at the 10-edit budget
pulling the mean well below the median. Extending the edit budget beyond
five rounds does not systematically improve either weak-start recovery or
baseline refinement.

\section{Discussion}
\label{sec:discussion}

\textsc{OptiArena} positions algorithm optimization between static code
generation and open-ended discovery: static benchmarks ask whether a model
writes code that passes tests, and discovery systems show that evaluator-guided
search can produce impressive artifacts. \textsc{OptiArena} targets the middle
case. Under a small fixed budget, can a model use feedback to improve an
executable algorithm in a way that survives held-out evaluation?

\noindent\textbf{From correctness to improvement.}
Starting from runnable policies rather than blank prompts changes what success
means: not syntactic validity or task completion, but behavioral improvement
over a known starting point under the same resource policy, a practical skill in
LLM-assisted workflows that edit imperfect artifacts after concrete feedback.

\noindent\textbf{Degradation is a first-class signal.}
Failures are not only invalid programs: a model can produce plausible code that
removes invariants, times out from over-expanded search, or swaps a strong
heuristic for a familiar weaker template. Treating degradation as an outcome
rather than noise lets \textsc{OptiArena} separate controlled local edits from
recognizable-but-weaker rewrites.

\noindent\textbf{Contamination is measured, not dismissed.}
Classic game algorithms are widely documented, so a game-based benchmark cannot
claim to be contamination-free; the question is how much measured optimization
depends on recognizable surface cues rather than executable improvement. We therefore
interpret the matched vanilla--obfuscated WSO comparison on the registered Cap
scale, without extending it to SBR.
The matched control is small in aggregate under both the raw Cap
difference and scale-normalized summaries, and neither aggregate interval
excludes zero. The game-level variation cautions against treating robustness to
surface obfuscation as a single model-wide property.

\noindent\textbf{Why games are useful despite not being the end domain.}
Games provide exact rules, controlled repeated execution, strong references, and
interpretable failure modes, a combination that is hard to obtain in realistic
software-engineering or discovery tasks. Their role here is diagnostic: a compact
substrate for studying how LLMs edit algorithms under feedback, with a policy
interface and archive that could be adapted to non-game executable tasks.

\noindent\textbf{Scope of the empirical claims.}
The results are bounded-resource optimization claims, not state-of-the-art game
play or autonomous discovery: a positive WSO result indicates recovered capability from a designated weak starter, a positive SBR result indicates local improvement over an editable competent baseline, and a small contamination shift would indicate robustness to surface
obfuscation. The contribution is that these claims are separated, measured, and
made auditable in one benchmark.

\section{Conclusion}
\label{sec:conclusion}

\textsc{OptiArena} reframes LLM evaluation around bounded-resource algorithm
optimization: rather than asking only whether a model can answer a static
question or synthesize a program once, it asks whether a model can edit
executable strategy code over feedback rounds under sealed references, fixed
budgets, and reproducibility gates, providing a reproducible, budget-explicit
foundation for studying LLMs as research tools in algorithmic workflows.
Empirically, the five-edit matrices show substantial but uneven recovery from
designated weak starters, while refinement of competent baselines is less
reliable and can be destructive; this contrast across models, games, and
reference signals shows that bounded-feedback optimization is a distinct
capability rather than a consequence of executable code alone.

\section*{Limitations}
\label{sec:limitations}

\textsc{OptiArena} studies bounded, feedback-driven code optimization in a
compact set of game environments. Its conclusions therefore concern relative
improvement under registered budgets and sealed references, rather than all
forms of programming or open-ended scientific discovery. Generality depends on
the selected games, references, splits, and model versions; the benchmark
mitigates these factors through held-out and stress evaluation, contamination
controls, archived candidates, and explicit run metadata.
The main matrices further use a single prompt condition, seed, and
five-edit budget with designated starter policies, so they do not establish
stability across prompts, seeds, longer search horizons, or alternative starter
strengths; the obfuscation control likewise covers only WSO surface
familiarity rather than semantic transfer, and some two-player records support
aggregate rather than trial-level paired inference.

\section*{Ethical Considerations}
\label{sec:ethics}

This work uses public game environments and does not involve human subjects or
personal data. The main risk is overclaiming the reliability of automatically
generated code or overstating the autonomy of LLM-driven research systems. We
mitigate this by running generated code only in a sandbox with disabled network
access, fixed resource limits, static validation, and sealed evaluators. The
candidate programs are benchmark artifacts and should not be treated as trusted
software. Artifact release must also respect the licenses of external engines
and solvers, including components that cannot be redistributed directly.
AI writing assistants were used to polish prose and grammar in this
paper, and AI coding assistants were used for parts of the benchmark's
implementation code. The experimental framework, benchmark design, and
reported experiments were conceived, implemented, and executed by the
authors.

\bibliography{ref}

\appendix
\section*{Appendix}
\startappendixcontents

This appendix is organized to make the benchmark auditable without interrupting
the main narrative. Appendix~\ref{app:task-registry} documents the task
registry and policy interface. Appendix~\ref{app:optimization-loop} specifies
the optimization loop, feedback schema, and prompt templates.
Appendix~\ref{app:metrics-details} defines metric fallbacks and result
surfaces. Appendix~\ref{app:experimental-protocol} records the experimental
design and reporting rules. Appendix~\ref{app:artifacts} describes artifact
release, qualitative case selection, and dependency provenance.

\apptableofcontents
\bigskip

\section{Task Registry and Environment Details}
\label{app:task-registry}

Table~\ref{tab:task-registry} summarizes the five environments in terms of the
algorithmic behavior they stress. The benchmark keeps a separation between
\emph{editable} policies and \emph{sealed} references: WSO and SBR starters are
source files exposed to the optimizer, whereas references are hidden strength
anchors used for calibration and final evaluation.

\begin{table}[!ht]
\centering
\small
\setlength{\tabcolsep}{4pt}
\begin{tabular}{@{}L{0.30\columnwidth}L{0.60\columnwidth}@{}}
\toprule
Task & Role in the suite \\
\midrule
Breakthrough 8$\times$8 & Race-and-capture lookahead. \\
Othello 8$\times$8 & Positional feature weighting. \\
Hex 11$\times$11 & Connectivity planning. \\
Connect 4 7$\times$6 & Saturated tactical search with solver-derived oracle diagnostics. \\
2048 & Stochastic value search and tile-success distributions. \\
\bottomrule
\end{tabular}
\caption{\textbf{Task suite summary.} The ``role'' column names the algorithmic
skill each game stresses. Exact engine versions, opening sets, resource budgets,
and sealed-reference identities are stored in artifact metadata.}
\label{tab:task-registry}
\end{table}

The starter/reference pairings are chosen to separate editable code from the
hidden strength anchor. Breakthrough pairs shallow tactical starters and
editable alpha-beta variants with a sealed alpha-beta reference. Othello uses
greedy or feature-based starters and evaluates against an Edax-backed
reference. Hex compares connectivity heuristics and editable search policies
against a MoHex-style reference. Connect 4 exposes tactical search starters
while retaining solver-derived oracle diagnostics for final analysis. The 2048
tasks pair simple corner or expectimax-style policies with stronger reference
policies used for calibration and stress evaluation. The released metadata
records the exact source hashes, engine ids, split ids, and budget ids for each
pairing.

Table~\ref{tab:anchor-construction} makes this pairing concrete: for
each game it names the designated weak starter and editable competent
baseline algorithms and reports their calibration performance against the
same fixed-budget opponent used in the GCR denominator
(\eqref{eq:gcr}, Sec.~\ref{sec:metrics}). This calibration opponent is a bounded-depth or
bounded-search dev-level reference, not the stronger sealed oracle
(BitBully/Edax/MoHex) used for the primary Cap estimator on Connect~4,
Othello, and Hex; the near-zero calibration span for Othello and Hex is
exactly why Appendix~\ref{app:saturated-games} registers a move-agreement
estimator for those two games instead of this win-rate span.
\begin{table*}[!ht]
\centering
\small
\setlength{\tabcolsep}{4pt}
\begin{tabular}{@{}L{0.13\textwidth}L{0.26\textwidth}L{0.37\textwidth}cc@{}}
\toprule
Game & Weak starter & Competent baseline & Naive & Comp. \\
\midrule
Breakthrough & Material/advancement greedy (1-ply) & Iterative-deepening negamax, alpha-beta, transposition table & 0.00 & 0.12 \\
Othello & Greedy disc-flip count & Native bitboard iterative-deepening alpha-beta, transposition table & 0.00 & 0.00 \\
Hex & Shortest-path connection cost (no search) & Threat/bridge-pattern lookahead over top candidate moves & 0.00 & 0.00 \\
Connect~4 & Plain minimax, center-column heuristic & Iterative-deepening alpha-beta, transposition table, window heuristic & 0.06 & 0.18 \\
2048 & Fixed directional preference (no search) & Iterative-deepening expectimax, transposition table, snake/merge heuristic & 7.80 & 9.35 \\
\bottomrule
\end{tabular}
\caption{\textbf{Anchor construction and calibration span.} Naive and Comp.\
are the registered weak-starter and competent-baseline scores against the
calibration opponent (win rate for Breakthrough/Othello/Hex/Connect~4, robust
score for 2048; 2048's naive score is the mean over the twelve models' own
iteration-0 evaluations, since it is stochastic). These are the
\(P_g(\pi_0^g)\) and \(P_g(\pi_{\mathrm{comp}}^g)\) terms in \eqref{eq:gcr}; a near-zero
gap (Othello, Hex) means the win-rate GCR is uninformative for that game, not
that the competent baseline lacks headroom under the sealed oracle.}
\label{tab:anchor-construction}
\end{table*}

\subsection{Policy Contract}
\label{app:policy-contract}

Every candidate source file exposes a single policy entry point. The callable
receives the current state, the legal-action list, the player identifier, and a
time budget, and it must return exactly one legal action. The harness owns rule
execution, legal-action generation, state encoding, timeout enforcement, and
result logging. Candidate policies may use the Python standard library plus the
libraries explicitly allowed in the prompt; they may not use network,
filesystem, subprocess, or project-internal imports. This contract is kept
intentionally small so that improvements are attributable to the policy code
rather than to privileged access to the evaluator.

\subsection{Native Implementations and Conformance}
\label{app:variant-conformance}

OpenSpiel is the rules authority wherever it provides the needed game and
parameterization. A few environments require native repository implementations,
most notably 2048 because OpenSpiel does not include it. These implementations
are admitted only after conformance checks against a reference implementation of
the same rules, covering legal-action generation, state transitions, terminal
detection, and scoring where applicable. Each environment defines disjoint
development, held-out, and stress openings or seed pools so that final evaluation
never reuses development material.

\section{Optimization Protocol and Prompting}
\label{app:optimization-loop}

Each evaluation instance begins with iteration 0, the unmodified starter. At
later iterations, the harness selects a parent from the archive, builds a prompt
from the parent source and prior feedback, asks the LLM for a complete
replacement source file, validates the candidate, and evaluates it in the
sandbox. The configuration uses epsilon-greedy parent selection over
the archive, with occasional exploration to avoid repeatedly editing a single
early winner.

\begin{enumerate}[leftmargin=*]
    \item \textbf{Initialize.} Load the WSO or SBR starter, run the registered
    initial evaluation, and store the result in the archive.
    \item \textbf{Select parent.} Choose an archived candidate according to the
    configured selection policy.
    \item \textbf{Prompt.} Provide the source, task rules, state encoding,
    compact feedback, recent history, and optional traces to the LLM.
    \item \textbf{Validate.} Extract the fenced Python source, check the
    required interface, and reject programs that violate static constraints.
    \item \textbf{Evaluate.} Run the candidate in the sandbox and record
    performance, failures, timing, and trace summaries.
    \item \textbf{Select final candidate.} After the edit budget is exhausted,
    choose the best development candidate and evaluate it on held-out and
    stress splits.
\end{enumerate}

The archive stores source files, parent relations, evaluation payloads, token
cost, latency, optional rationale text, and diff summaries. These records
support cost-efficiency curves, degradation analysis, edit-taxonomy summaries,
and post-hoc audits.

\subsection{Feedback Schema}
\label{app:feedback-schema}

The evaluator records both successful candidates and failures, but the optimizer
receives only a compact summary. Feedback is grouped into four families:
validity outcomes (syntax, interface, illegal moves, crashes, and timeouts),
performance outcomes (win rate, score distribution, confidence intervals, and
task-specific robust scores), efficiency outcomes (game length and move-time
statistics), and optional trace summaries for representative games. The LLM
does not see sealed-reference source code, hidden openings, solver values, or
the full final-evaluation set. Exact JSON field names are part of the released
artifact schema rather than the paper narrative.

\subsection{Prompt Templates}
\label{app:prompts}

The prompt is assembled from a fixed system contract and a fixed sequence of
user-message fields; only the requested reasoning style varies across
conditions. The skeleton is reproduced schematically below (exact wording ships
with the artifacts):

\begin{quote}\small
\textbf{System.} Return one complete policy source file; keep the required entry
point (a single callable taking the state, legal actions, player id, and time
budget, and returning one legal action); import only allowed libraries; do not
access evaluator internals, the network, or the filesystem; respect the per-move
time and memory budgets.

\textbf{User.} iteration id $\rightarrow$ hidden-reference notice $\rightarrow$
task objective $\rightarrow$ game rules $\rightarrow$ state encoding $\rightarrow$
parent source $\rightarrow$ compact feedback (validity, performance, efficiency)
$\rightarrow$ recent history $\rightarrow$ condition-specific instructions.

\textbf{Output.} A single fenced \texttt{python} code block; any other output is
recorded as an invalid candidate.
\end{quote}

\subsubsection{Shared System Instructions}
\label{app:system-prompt}

All prompt conditions share the same system-level contract. The system message
asks the model to return one complete policy source file, preserve the required
policy entry point, use only the allowed libraries, avoid evaluator internals
or external tools, and respect the move-time and memory budgets. The condition
only changes the requested amount of rationale or trace reading. Exact prompt
text is released with the artifacts; the paper reports the shared constraints
rather than reproducing implementation prompts.

\subsubsection{User Message Fields}
\label{app:user-prompt}

The user message is assembled from a fixed sequence of fields: iteration id,
hidden-reference notice, task objective, game rules, state encoding, parent
source, compact feedback, recent history, and condition-specific instructions.
Keeping this order fixed lets prompt-condition ablations change the requested
reasoning style without changing task information.

\subsubsection{Prompt Conditions}
\label{app:prompt-conditions}

\begin{table}[!ht]
\centering
\small
\setlength{\tabcolsep}{4pt}
\begin{tabular}{@{}L{0.24\columnwidth}L{0.62\columnwidth}@{}}
\toprule
Condition & Instructional difference \\
\midrule
\texttt{code\_only} & Scalar feedback only; output only the Python code block. \\
\texttt{rationale} & Add a short rationale before the code block. \\
\texttt{reflection} & Add failure analysis and a fix plan before the code block. \\
\texttt{trace\_aware} & Read the supplied sample traces, then write a fix plan and code. \\
\bottomrule
\end{tabular}
\caption{\textbf{Prompt-ablation conditions} supported by the harness. The
primary comparison uses \texttt{trace\_aware}; other conditions are reported as
appendix ablations when included.}
\label{tab:prompt-conditions}
\end{table}

\section{Metrics and Result Surfaces}
\label{app:metrics-details}

\subsection{Metric Details for Saturated Games}
\label{app:saturated-games}

Some games are intentionally evaluated against very strong references. In those
settings, raw win rate can saturate at zero for both the designated weak starter and the editable competent baseline,
which makes a win-rate GCR uninformative. \textsc{OptiArena} therefore registers
a per-game capability estimator in advance, chosen by the strongest reference
signal each game exposes, and discloses it in the table footnote:
\begin{itemize}
\item \textbf{Solved-game oracle} (Connect~4 / BitBully): the move-optimal rate,
i.e.\ the fraction of candidate moves in the oracle arg-max set, with mean regret
as an additional diagnostic. Move-level optimality is used here in preference to
a win-rate GCR because it scores move quality directly even when win rate is
saturated.
\item \textbf{Policy reference} exposing only a chosen action (Hex / MoHex,
Othello / Edax): the move-agreement rate with the reference move.
\item \textbf{Single-player stochastic} (2048): a reference-\(\Delta\), the
candidate robust score normalized to the weak-starter-to-reference span,
\((R_T-R_0)/(R_{\mathrm{ref}}-R_0)\).
\item \textbf{Win-rate gap} (Breakthrough): GCR over the calibrated
weak-starter-to-competent-baseline gap, used where the gap is well conditioned.
\end{itemize}
The move-level metrics are computed after evaluation by replaying stored traces
and never expose oracle values to the LLM during optimization.

The Cap/\(\Delta\) table separates the selected per-game capability
estimator (Cap) from the sign of its change from the starter to the best
candidate (\(\Delta\)) on the same capability scale, keeping saturated games in
the benchmark while making the estimator used for each game explicit.
Validity, the completed-trial rate, is near-constant across cells and is
therefore reported only where it falls below 1.00.

\subsection{Metric Normalization and Diagnostics}
\label{app:metric-diagnostics}

Contamination sensitivity is reported both as a signed difference and, when
useful, as a scale-normalized quantity. For a condition \(c\), the signed score
is
\[
\mathrm{CSI}_{c}(g,m)=M_{\mathrm{vanilla}}(g,m)-M_{c}(g,m),
\]
where \(M\) is the registered per-game WSO capability estimator
(Appendix~\ref{app:saturated-games}). The normalized form divides
this difference by
\[
\max(|M_{\mathrm{vanilla}}(g,m)|, |M_c(g,m)|,\epsilon_{\mathrm{floor}}),
\]
which avoids overstating ratios when both metrics are close to zero. A positive
value indicates an advantage for the vanilla surface, while values near zero
indicate similar behavior under the comparison condition.

For obfuscation, game dynamics are preserved, so CSI is interpreted as
sensitivity to surface familiarity rather than to any change in task difficulty.
This control is scoped to WSO; no obfuscated SBR comparison is inferred from it.
We aggregate over the 60 matched model--game cells with 10,000 paired
bootstrap resamples. The mean CSI-Cap is $-0.043$ (95\% CI
$[-0.124, 0.043]$); applying the normalized form above before aggregation gives
$0.016$ (95\% CI $[-0.105, 0.144]$).

Robustness diagnostics are computed from the same selected candidate. The
development-to-held-out gap measures whether the optimization loop overfits to
the feedback split; the held-out-to-stress gap measures whether the candidate
remains stable under the registered harder distribution. Cost is computed from
LLM usage metadata rather than local game-evaluation wall-clock. Degradation
rate is the fraction of model-game entries in which the selected candidate is
worse than its starter on the registered capability scale.

All aggregate tables require matching calibration rows, split ids, reference
identities, resource policies, and final-evaluation records. Missing required
inputs are not converted into default scores or silently omitted from
aggregates.

\subsection{Result Tables and Provenance}
\label{app:appendix-result-tables}

The per-game WSO and SBR matrices are reported in the main text
(Tables~\ref{tab:wso-full} and~\ref{tab:sbr-full}). Their entries are populated
from archived SQLite databases and analysis JSON so that every value remains
tied to recorded evaluation metadata.

\subsubsection{Diagnostic Surfaces}
\label{app:diagnostic-analyses}

Beyond the aggregate tables, the analysis scripts compute three diagnostic
surfaces directly from stored runs: per-model per-game WSO CSI between the vanilla and
obfuscated conditions; validity and degradation summaries that expose gains
driven by brittle rewrites, invalid programs, or excessive search; and
cost-efficiency and edit-type summaries relating improvement to optimization
effort. These are released with the artifact; the main text reports the WSO and
SBR matrices and the matched WSO contamination summary in Table~\ref{tab:rq3-csi}.

\section{Experimental Design and Reporting Rules}
\label{app:experimental-protocol}

This appendix fixes the reporting scope and inclusion rules used for benchmark
comparisons, separating confirmatory matrices from pilot runs and ablations.

\subsection{Reporting Scope}
\label{app:reporting-scope}

The empirical design separates two primary matrices and one control: WSO on the vanilla tasks, matched
SBR on the same tasks and model roster, and a WSO contamination control that
compares vanilla tasks with obfuscated counterparts. Each comparison is reported
only when its coverage and metadata checks pass.
Development pilots are used only to validate the harness and are excluded from
result tables. Additional random seeds and prompt ablations are treated as
stability analyses. Aggregate claims are made only for a complete
matrix over models and games under the stated comparison; partial control runs are
reported separately and are not mixed into the main WSO or SBR averages.
A supplementary edit-budget extension (Section~\ref{sec:diagnostic-analysis})
explores longer optimization horizons on a subset of games; it sits outside
this confirmatory scope and is not treated as a fourth matrix.

\subsection{Model Roster}
\label{app:model-roster}

Table~\ref{tab:model-roster} groups the model roster used for cross-model
comparison. Closed models serve as high-capability and lower-latency proprietary
comparison points; open-weight models test whether the benchmark is informative
for systems with different cost and scale profiles. Provider ids are recorded in
the artifact metadata.

\begin{table}[!ht]
\centering
\footnotesize
\setlength{\tabcolsep}{4pt}
\begin{tabular}{@{}L{0.32\columnwidth}L{0.58\columnwidth}@{}}
\toprule
Group & Models \\
\midrule
Closed-source frontier & Claude Sonnet 4.6; GPT-5.4; GPT-5.1-Codex \\
Closed-source fast & Gemini 3.5 Flash \\
Open reasoning/coding & DeepSeek-V4-Pro; Qwen3-Coder; Kimi-K2.6 \\
Open smaller/low-cost & Qwen3-14B; Gemma-4-26B; MiniMax-M2.7; Mistral-Small; GLM-4.7 \\
\bottomrule
\end{tabular}
\caption{\textbf{Model roster.} A finer grouping of the twelve models than the
proprietary/open-weight split used in the main text
(Table~\ref{tab:rq-optimization}). Exact provider ids and version strings are
stored with the run metadata.}
\label{tab:model-roster}
\end{table}

\subsection{Run Budget}
\label{app:run-config}

The primary run gives each model five edit calls after iteration 0 evaluates the unmodified starter policy.
Intermediate candidates are evaluated on a small development sample to guide the
optimization loop; the selected candidate is then evaluated on development,
held-out, and stress splits with \(n=50\) games or trials per split. Two-player
policies receive a one-second per-move budget, while 2048 uses a 0.1-second
budget to keep stochastic rollouts inexpensive. Candidates run in a sandboxed
Docker evaluator with one CPU, 2048 MB memory, process limits, and disabled
network access. The primary prompt condition is trace-aware; prompt ablations
use the same task, split, and resource policy.

This budget makes the unit cost explicit. A single cell for a model, game,
regime, and surface uses at most five LLM edit calls after the starter and a fixed number of
development, held-out, and stress evaluations. The main vanilla WSO/SBR matrix
therefore contains 12 models, five games, and two regimes before surface
controls. LLM token usage, provider-reported cost, local evaluator wall-clock,
and invalid-candidate counts are logged separately so that optimization gains
can be read against the amount of computation spent to obtain them.
Across the three vanilla WSO, SBR, and obfuscated-WSO
matrices, the 540 development, held-out, and stress final-evaluation records
(180 per matrix) total 343.9 local evaluator wall-clock hours (113.5 for WSO,
129.8 for SBR, and 100.5 for the obfuscated-WSO control), recorded independently
of the LLM cost reported in Table~\ref{tab:rq-optimization}.

\subsection{Inclusion and Rerun Rules}
\label{app:inclusion-rules}

An evaluation instance may be rerun only for infrastructure failures such as
evaluator crashes, external-engine misconfiguration, Docker failures, storage
failures, or transient API errors. A low score, destructive edit, timeout
caused by the candidate code, or invalid model-generated program is a valid
benchmark outcome and is not a rerun reason. If an entire game must be excluded
because of an infrastructure defect, it is excluded for all models in the
affected comparison and the exclusion is reported explicitly.

\subsection{Reproducibility Checks}
\label{app:reproducibility-gates}

A run is included in reported aggregates only if every evaluation instance
satisfies the same metadata checks. The rules implementation, starter source,
sealed reference id, reference budget, split id, sandbox image, memory and
process limits, per-move budget, model id, prompt condition, and random seed
must be recorded. Development, held-out, and stress openings or seed pools must
be disjoint and replayable. Native implementations must pass conformance tests against a reference
implementation of the same rules before they are used.

Calibration rows are required for the designated weak starter, editable competent baseline,
and SBR empirical baseline. These rows must use the same reference identity
and resource policy as the final sweep, with sample size and confidence
intervals recorded. The analysis requires these inputs rather than substituting
default values. In particular, missing calibration, missing final evaluations,
reference-id mismatches, or absent held-out/stress rows are excluded from
aggregate reporting rather than converted into defaults.

Uncertainty estimates are generated from stored final-evaluation rows. For
two-player games, intervals are computed over paired openings and side
assignments where available; for stochastic tasks, they are computed over seed
pools. Cross-model summaries keep model-game cells as the unit of aggregation
so that a large number of trials from one game does not dominate the reported
comparison.

\section{Qualitative Cases and Artifacts}
\label{app:artifacts}

\subsection{Qualitative Audit Samples}
\label{app:qualitative}

The archive supports qualitative auditing through cases selected by explicit
criteria rather than narrative convenience, spanning a successful weak-start recovery, a destructive SBR edit, an invalid-code failure, and a timeout caused by
excessive search. Each sampled case is stored with its run id,
evaluation-instance id, iteration, split, and selection criterion, so that the
source diffs and evaluator traces behind the quantitative patterns can be
replayed from the released artifact.

\begin{table}[h]
\centering
\footnotesize
\setlength{\tabcolsep}{3pt}
\begin{tabular}{@{}L{0.2\textwidth}L{0.22\textwidth}}
\toprule
Case (cell id) & Selection criterion \\
\midrule
MiniMax-M2.7, 2048, Track A, iteration 2 (\texttt{397a8655d077ace0}) &
Largest Cap gain among algorithm-inserted WSO cells, single-player games \\
Gemma-4-26B, Breakthrough, Track A, iteration 5 (\texttt{093d3411facd7944}) &
Largest Cap gain among algorithm-inserted WSO cells, two-player games \\
GLM-4.7, Connect~4, Track B, iteration 1 (\texttt{da06291748c5db45}) &
Most negative $\Delta_{\mathrm{SBR}}$ among destructive-edit SBR cells \\
DeepSeek-V4-Pro, 2048, Track B, iteration 5 (\texttt{2280c6ab272a87a4}) &
Most positive $\Delta_{\mathrm{SBR}}$ among positive-edit SBR cells \\
Gemma-4-26B, Connect~4, Track A, iteration 2, dev quick-eval (\texttt{ccdf097934f73b8a}) &
Sole quick-eval compile failure archived under Track A \\
Claude Sonnet 4.6, 2048, Track A, iteration 4, dev finalize-eval (\texttt{16567b2203766cad}) &
Sole finalize-stage candidate runtime failure (container OOM under excessive search) archived under Track A \\
\bottomrule
\end{tabular}
\caption{Rule-selected qualitative audit cases. The first four illustrate the
edit-type census in Section~\ref{sec:diagnostic-analysis}; the last two are
the WSO archive's only instances of an invalid-code failure and a
resource-exhaustion failure from excessive search, respectively, and are
included as-is rather than chosen from a larger pool.}
\label{tab:qualitative-cases}
\end{table}
The first two rows keep the starter's underlying interface but replace its
decision procedure: MiniMax-M2.7's candidate adds a depth-2 expectimax that
scores both tile-spawn outcomes (probability 0.9 for a 2-tile, 0.1 for a
4-tile) before re-evaluating the board, and Gemma-4-26B's candidate adds a
2-ply minimax search with alpha-beta pruning over a material-and-advancement
heuristic. The third row rewrites the competent baseline's own
iterative-deepening alpha-beta search from scratch behind a new
numpy-based board representation rather than editing it incrementally; the
rewrite's own comments flag uncertainty about an action encoding inherited
from the parent ("assuming legal actions are strings..."), a plausible source
of the resulting regression. The fourth row instead edits the existing
evaluation function in place, folding the immediate merge score into the
search's leaf value and retuning heuristic weights (lowering the empty-cell
weight from 270 to 10 while raising the corner and monotonicity weights)
rather than replacing the search. The fifth row failed to import as a Python
agent during the three-game quick evaluation and was retained in the
optimization archive rather than silently repaired or excluded. The sixth
row's container was OOM-killed during the dev finalize evaluation,
contributing to that cell's sub-1.00 validity reported in
Table~\ref{tab:wso-full}.

\subsection{Archived Artifacts}
\label{app:archived-artifacts}

The artifact archive contains the run database, generated analysis tables,
candidate sources, configuration files, and calibration records. The database
stores cells, iterations, final evaluations, budgets, and reference baselines;
the source archive stores every starter and generated candidate; and the
analysis outputs include JSON/CSV summaries, cost records, robustness tables,
the WSO CSI table, Val/Cap/\(\Delta\) tables, and qualitative-audit samples. The
release also includes calibration sidecars and the experiment definitions needed
to replay each reported comparison.

\subsection{External Dependency Notes}
\label{app:external-dependencies}

The vanilla rules are backed by OpenSpiel where available. Connect 4 uses a
BitBully-backed reference/oracle, Othello uses Edax, Hex uses MoHex, and 2048
uses a nneonneo-style reference policy. Artifact metadata reports versions,
engine ids, budgets, and license constraints. GPL or AGPL engines are not
bundled directly when their licenses prohibit redistribution in the main
artifact package.

For reproducibility, the artifact records the rules authority, sealed reference
identity, sandbox image, calibration rows, split identifiers, analysis script
version, and token-cost records for every reported comparison. These metadata are
checked before aggregate tables are generated.

\end{document}